\documentclass{article}
\usepackage{spconf,amsmath,graphicx,hyperref}
\usepackage{subcaption} % in preamble
\usepackage{verbatim} % in preamble
\usepackage{balance}
\usepackage{multirow}
\usepackage{graphicx}
\usepackage{amsmath}
\usepackage{hyperref}
\title{Prediction and Causality of functional MRI and synthetic signal \\using a Zero-Shot Time-Series Foundation Model }
\name{Alessandro Crimi$^{1}$ \qquad \qquad Andrea Brovelli$^{2}$ \thanks{Send correspondence to alecrimi@agh.edu.pl}}
\address{
$^{1}$AGH University of Krakow, Poland\\
$^{2}$Institut de Neurosciences de la Timone UMR 7289, Aix Marseille Université, CNRS, 13005, Marseille, France
}
\begin{document}
%\ninept
%
\maketitle
\begin{abstract}
% Max 200 words
Time-series forecasting and causal discovery are central in neuroscience, as predicting brain activity and identifying causal relationships between neural populations and circuits can shed light on the mechanisms underlying cognition and disease. With the rise of foundation models, an open question is how they compare to traditional methods for brain signal forecasting and causality analysis, and whether they can be applied in a zero-shot setting.

In this work, we evaluate a foundation model against classical methods for inferring directional interactions from spontaneous brain activity measured with functional magnetic resonance imaging (fMRI) in humans. Traditional approaches often rely on Wiener–Granger causality. We tested the forecasting ability of the foundation model in both zero-shot and fine-tuned settings, and assessed causality by comparing Granger-like estimates from the model with standard Granger causality. We validated the approach using synthetic time series generated from ground-truth causal models, including logistic map coupling and Ornstein–Uhlenbeck processes. The foundation model achieved competitive zero-shot forecasting fMRI time series (mean absolute percentage error of 0.55 in controls and 0.27 in patients). Although standard Granger causality did not show clear quantitative differences between models, the foundation model provided a more precise detection of causal interactions.

Overall, these findings suggest that foundation models offer versatility, strong zero-shot performance, and potential utility for forecasting and causal discovery in time-series data.
\end{abstract}
\begin{keywords}
Time series, Granger causality, fMRI, LLM, foundation models, ARIMA
\end{keywords}

\begin{figure} 
\includegraphics[width=\linewidth]{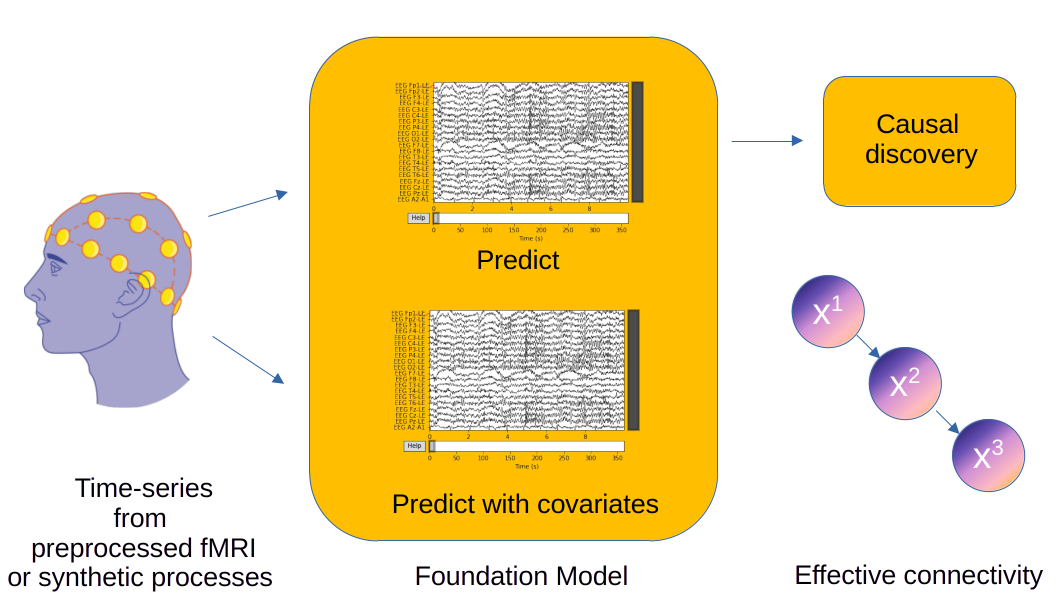}
\caption{Overview of the experiments: first we investigate the predictive power of the TimeSeries foundation model with brain signals, and then we evaluate if the time series predicted with it, can also be used for causal discovery.}
\label{fig:causality}
\end{figure}

\section{Introduction}
\label{sec:intro}
Time-series analysis in neuroscience is of considerable importance, as it enables the characterization of dynamic brain processes and the inference of underlying mechanisms; however, it remains challenging due to the high dimensionality, noise, and intrinsic complexity of neural signals \cite{biswal2010toward}.   
Time series are also the basis for network-level analyses of brain activity and inference of effective and functional connectivity \cite{friston2011functional}, quantifying relevant relationships and communication in the brain that can ultimately be exploited as biomarkers. Hence, accurate forecasting of neural time series is increasingly valuable for neuroimaging and neuroscience application. 
Recent advances in foundation models for time series, such as the Time series Foundation model (TimesFM) \cite{das2024decoder}, Time-Mixture of Expert (Time-MOE) \cite{shi2024time}, Llama-lag \cite{rasul2023lag}  and others, promise zero-shot forecasting capabilities that could revolutionize brain science. 
Analogous to how large language models represent words as embeddings, these models encode time series into latent embeddings and are designed to predict future trajectories over a specified horizon given an initial sequence as input. Importantly, they are developed for zero-shot use: pre-trained on large and diverse collections of time series data, they theoretically enable forecasting without task-specific training, thereby offering advantages such as broad applicability and reduced reliance on domain-specific datasets. 
Recent studies have introduced transformer-based models trained from scratch on electroencephalography (EEG) data \cite{chen2024eegformer,chen2025large}, functional MRI (fMRI) \cite{wang2025towards}, or combined EEG–fMRI datasets \cite{bayazi2024general}. While these approaches have shown promising results, they face recurring challenges arising from data heterogeneity (e.g., differences in electrode counts, montages, sampling rates, scanner protocols) and inter-subject variability, which necessitate robust pre-training objectives and augmentation strategies. These considerations suggest that general-purpose foundation models may provide a more viable solution. Building on this perspective, our first goal is to evaluate whether the performance of a zero-shot, domain-agnostic model meaningfully differs from that of traditional statistical methods specifically developed for brain data.

Among traditional approaches to signal prediction, autoregressive integrated moving average (ARIMA) models \cite{box2015time} remain the most widely used and have demonstrated robust performance in neuroimaging applications, often outperforming neural network–based methods \cite{ganesan2021human}. Accordingly, we compare the predictions of the zero-shot model against ARIMA and related statistical techniques.

To date, most studies have focused exclusively on time-series forecasting, while extensions to causality and effective connectivity remain unexplored. In this work, we propose to investigate whether a foundation model can be adapted for causal inference by leveraging autocorrelates, in analogy to Granger causality, which remains the most widely used method for estimating directional interactions from neural time series. A central limitation of causal discovery in neuroscience is the absence of ground-truth in real data. To address this, we validate our approach using two well-characterized synthetic systems: coupled logistic maps and multivariate Ornstein–Uhlenbeck processes. These models not only provide explicit ground-truth, but also enable the distinction between excitatory and inhibitory causal interactions, which is particularly relevant in the context of brain signaling. For this evaluation, we selected TimesFM, as it natively supports time-varying covariates, a key requirement for our causality analysis that is not fulfilled by other models.

\vspace{-0.3cm}
\section{Methods}
\label{sec:methods}
\subsection{Dataset and Pre-processing}
We used three datasets in this study. The first two dataset are synthetic datasets designed to test causal discovery: one with causal relationships defined by coupled logistic maps, and the other based on multivariate Ornstein–Uhlenbeck processes. The third dataset is a real-world dataset comprising both healthy and patient participants, used to evaluate differences in the prediction of healthy versus pathological fMRI time series. 
\vspace{-0.1cm}
\subsubsection{Synthetic data}
We generate synthetic data sets with known ground-truth causality.  
\textbf{Logistic Map Coupling:} Three unidirectionally coupled time series $\{X^{(1)}_t\}$, $\{X^{(2)}_t\}$, and $\{X^{(3)}_t\}$ ($n = 100$) were generated with initial conditions $X^{(j)}_0 = c_j + \epsilon_j$ ($c_1=0.1, c_2=0.2, c_3=0.3$; $\epsilon_j \sim U(-0.01, 0.01)$). The update equations were as follows:
\[
\begin{aligned}
X^{(1)}_t &= r X^{(1)}_{t-1}(1 - X^{(1)}_{t-1}), \\
X^{(2)}_t &= r X^{(2)}_{t-1}(1 - X^{(2)}_{t-1}) + \alpha X^{(1)}_{t-1}, \\
X^{(3)}_t &= r X^{(3)}_{t-1}(1 - X^{(3)}_{t-1}) + \alpha X^{(2)}_{t-1},
\end{aligned}
\]
where $r$ and $\alpha$ are coupling coefficients. We generated 10 simulations with $\alpha$ ranging from 0.1 to 0.9. \textbf{Multivariate Ornstein-Uhlenbeck (MOU):} For the $N=10$ nodes, we simulated MOU processes governed by $d\mathbf{X}_t = C \mathbf{X}_t  dt + \Sigma^{1/2} d\mathbf{W}_t$, where $C$ is a random connectivity matrix with density $d \in (0,1)$ (nonzero entries uniformly sampled from $\left[ -\frac{1}{Nd},  \frac{1}{Nd} \right]$) and $\Sigma = \sigma^2 I_N$ ($\sigma^2 = 0.2$). We generate 10 networks for each density $d$ from 0.1 to 0.9. 

\subsubsection{Human fMRI data}
The neuroimaging data were previously acquired by the School of Medicine at Washington University in St. Louis, with full acquisition and clinical procedures described in \cite{corbetta2015common}. Briefly, the dataset includes 26 healthy control participants and 104 stroke patients who underwent fMRI scanning in the acute post-stroke phase. For the present study, we selected 26 control subjects and randomly sampled 26 stroke patients to obtain a balanced cohort. Preprocessing of the fMRI data was performed using fMRIPrep 23.1.3 \cite{esteban2019fmriprep}. %, built on Nipype 1.8.6 \cite{gorgolewski2011nipype}. 
The pipeline included skull stripping, spatial normalization to a standard brain template, and nuisance regression with 36 confounding parameters. The voxel-wise 4D signal was then parcellated into 117 regions of interest (ROIs) using the Schaefer atlas \cite{schaefer2018local}, yielding 117 regional time series per subject. Series were also MinMax scaled [0,1] prior to analysis. For each subject, 600 time points (20 minutes) were extracted and split into training (first 540 time points) and testing (remaining 60 time points) sets, corresponding to a 90\%–10\% split. %This design ensured sufficient data for model fitting while preserving a clinically relevant forecast horizon.

\subsection{Forecasting Models}
We compared TimesFM—a 200M-parameter pre-trained model, evaluated with default hyperparameters (batch size=32, GPU backend) against several baselines: i) naive forecasters (mean strategy $\hat{y}_{t+1} = \frac{1}{N} \sum{i=1}^N y_i$ and last-value strategy $\hat{y}_{t+1} = y_t$); ii) linear regression (LR) (window length=60); iii) ARIMA(p,d,q=5; no seasonality); iv) Error, Trend, and Seasonality (ETS) with automated trend and damping selection.   
% Se ci arrivo la includo, 
We acknowledge that a trained long short-term memory
(LSTM) network can represent a strong baseline comparison \cite{lotte2018review}. However, the primary objective of this work is to evaluate the zero-shot, out-of-the-box applicability of foundation models against classical statistical methods requiring minimal training that are the current standard in neuroimaging research.  It can be considered for future investigation in comparison to fine-tuning the zero-shot model. 
%\subsection{Evaluation Framework}
%All models were evaluated using three complementary metrics: Mean absolute error (MAE), mean absolute percentage error (MAPE), and mean absolute scaled error (MASE) defined as
All models were evaluated using the mean absolute percentage error (MAPE), defined as 
%\begin{equation}
%\text{MAE} = \frac{1}{n}\sum_{i=1}^{n}|y_i - \hat{y}_i|
%\end{equation}
$ \frac{100\%}{n}\sum_{i=1}^{n}\left|\frac{y_i - \hat{y}_i}{y_i}\right|$
%\begin{equation}
%\text{MASE} = \frac{\text{MAE}}{\frac{1}{T-1}\sum_{t=2}^T |y_t - %y_{t-1}|}
%\end{equation}
%Moreover, we keep track of the computational time on the same machine considering the training time and inference time per forecast (in milliseconds).  
\vspace{-0.3cm}
\subsection{Causality analysis}
Traditional approaches rely on the Wiener–Granger causality principle, which is based on predictability: if past values of one time series improve the prediction of another (beyond the latter’s past values alone), the first is said to Granger-cause the second \cite{granger1980, bressler2011}.  Granger causality can be computed by comparing a restricted autoregressive (AR) model of $Y$ against a full model that also includes lagged values of $X$. The significance of the improvement is tested using an F-test on the residual variances.  To expand this reasoning to time series modeled with a foundation model, we consider additional time series as covariates to build a full model and inspect the residuals.  Here, the foundation model also generates predictions $\hat{Y}_t$ based on historical data
$\hat{Y}_t = \text{TimesFM}(Y_{t-w:t-1})$, 
where $w$ is the window size (context length) and $Y_{t-w:t-1} = \{Y_{t-w}, Y_{t-w+1}, \ldots, Y_{t-1}\}$. The residuals between the observed and predicted data of the foundation model are: $r_t = Y_t - \hat{Y}_t$. To computed Granger causality from the foundation model, we tested whether lagged covariates explain the residuals that the foundation model cannot capture. 
In the reported synthetic experiments, we fixed the total length of the MOU time series to 100 time points. Thus, the context window for TimesFM was set to $w=30$, which represents one third of the total length of the series, constituting a sufficiently long interval. For a given lag $\ell$, we computed the Pearson correlation between residuals and lagged covariates:
$\rho_\ell = \text{corr}(r_{t+\ell}, X_t)$.  We also fit a linear regression model
$r_{t+\ell} = \delta + \theta_\ell X_t + \eta_t$, where $\delta$ is the intercept  and $\theta_\ell$ is the regression coefficient for the lagged covariate $X_t$.

The best value among the results between the interval between 1 and 5 was chosen after Benjamini-Hochberg false discovery rate correction. The coefficient of determination $R^2$ measures the proportion of residual variance explained by the lagged covariate $X_t$:

\begin{equation}
R^2 = 1 - \frac{SS_{res}}{SS_{tot}} = 1 - \frac{\sum_t \eta_t^2}{\sum_t (r_{t+\ell} - \bar{r})^2}
\end{equation}
For the correlation test, we test $H_0: \rho_\ell = 0$ using: $t = \rho_\ell \sqrt{\frac{n-2}{1-\rho_\ell^2}} \sim t_{n-2}$, 
where $n$ is the number of aligned samples.

In summary, for the \textbf{classical Granger test}, $X$ is said to Granger-cause $Y$ if the F-statistic is significant, indicating that including lagged values of $X$ significantly improves the prediction of $Y$. In contrast, under the \textbf{TimesFM residual method}, $X$ is considered to have a causal influence on $Y$ if lagged values of $X$ are significantly correlated with, or explain a significant portion of, the residuals $r_t$—i.e., the component of $Y$ not captured by the foundation model. This indicates that $X$ contains predictive information about $Y$ beyond what TimesFM could account for. In practice, directionality is deemed significant at a threshold of $\alpha = 0.05$, correcting for multiple testing if we choose among many lags. 

\begin{comment}
\begin{figure*}[!h]
\centering
\begin{subfigure}[b]{0.45\linewidth}
    \includegraphics[width=\linewidth]{control.png}
   \caption{ } \vspace{-0.3cm}
    \label{fig:forecast}
\end{subfigure}
\hspace{0.05\linewidth} % spacing between images
\begin{subfigure}[b]{0.45\linewidth}
    \includegraphics[width=\linewidth]{patient.png}
 \caption{} \vspace{-0.3cm}
    \label{fig:causality}beginning
\end{subfigure}
\caption{Average MAPE across all brain regions and for all (a) control subjects, and (b) for all stroke patients.}
\label{fig:results}
\end{figure*}
\end{comment}

The causal discovery was evaluated by computing the mismatch in directionality, whether a true  causality was detected or not, and whether it was for example $X^{(1)} \rightarrow X^{(2)}$ or vice-versa. We tested mostly the synthetic data, as even if reported the average total causality for the fMRI data, we cannot evaluate it against a ground-truth. For the logistic coupling accuracy, precision and recall are sufficient. For the MOU, we need to take into account the sign of causality (excitatory or inhibitory). The foundation model used was TimesFM 1.2.0 accessed using Python API 3.11 and PyMOU to generate the MOU processes \cite{gilson2016estimation}. The code is accessible at URL \url{https://github.com/alecrimi/timesFM_stroke}. 

\vspace{-0.3cm}
\section{Results}
\label{sec:results}
\vspace{-0.2cm}
%\subsection{Forecasting Performance}
Table \ref{tab:performance}   summarizes the precision of the forecast between the methods and the subject groups. We quantified model forecasting performance using the mean absolute percentage error (MAPE) detailed above. TimesFm produced lower MAPE. However, assuming non-paired data across the brain regions, we found non-significant the results for the control subjects, and only significant  ($pval<0.05$) the results for the patient dataset. 
%Table \ref{tab:performancepatient} reports how the forecast metrics improve in percentage if we perform fine-tuning.  
Fine-tuning the TimesFM model led to an improvement 8\% for control subjects and 14\% for stroke patients, quantified as
$ 0.50 \pm 0.17$ for control and $ 0.23 \pm 0.01$ for stroke patients.  Regarding the causality analysis, the quantitative error was for the 3-node synthetic data is reported in Table \ref{tab:3node}

\begin{table}[!ht]
\centering
\caption{Forecasting Performance (Mean MAPE $\pm$ Variance)}
\begin{tabular}{lcc}
\hline
\textbf{Method} & \textbf{Control} & \textbf{Patient} \\
\hline
TimesFM zero-shot     & 0.55 $\pm$ 0.42 & 0.27 $\pm$ 0.01 \\
LR           & 0.59 $\pm$ 0.51 & 0.39 $\pm$ 0.01 \\
Naive Mean   & 0.57 $\pm$ 0.46 & 0.32 $\pm$ 0.01 \\
Naive Last   & 0.59 $\pm$ 0.37 & 0.32 $\pm$ 0.02 \\
ARIMA        & 0.61 $\pm$ 0.64 & 0.35 $\pm$ 0.04 \\
ETS          & 0.63 $\pm$ 0.51 & 0.32 $\pm$ 0.02 \\
\hline
\end{tabular}
\label{tab:performance}
\end{table}

\begin{comment}
\begin{table}[ht]
\centering
\caption{Increase in performance with fine-tuning}
\begin{tabular}{lcc}
\hline
\textbf{Metric} & \textbf{Control (\%)} & \textbf{Patient (\%)} \\
\hline
MAE   & 8.6  & 12.0 \\
MAPE  & 8.0  & 14.0 \\
MASE  & 8.8  & 15.7 \\
\hline
\end{tabular} \label{finetuning}
\end{table}
\end{comment}
%\subsection{Computational Requirements}
%Figure \ref{fig:compute} shows the trade-off between accuracy and computational resources. 
%Regarding the computational costs, it was observed that while TimesFM requires no training time (zero-shot), its inference time is nearly 10× higher than ARIMA. Memory usage follows a similar pattern, with TimesFM requiring 2.1GB versus ARIMA's 350MB.
\begin{figure*}[!t]
  \centering
  \begin{subfigure}[b]{0.25\textwidth}
    \includegraphics[width=\linewidth]{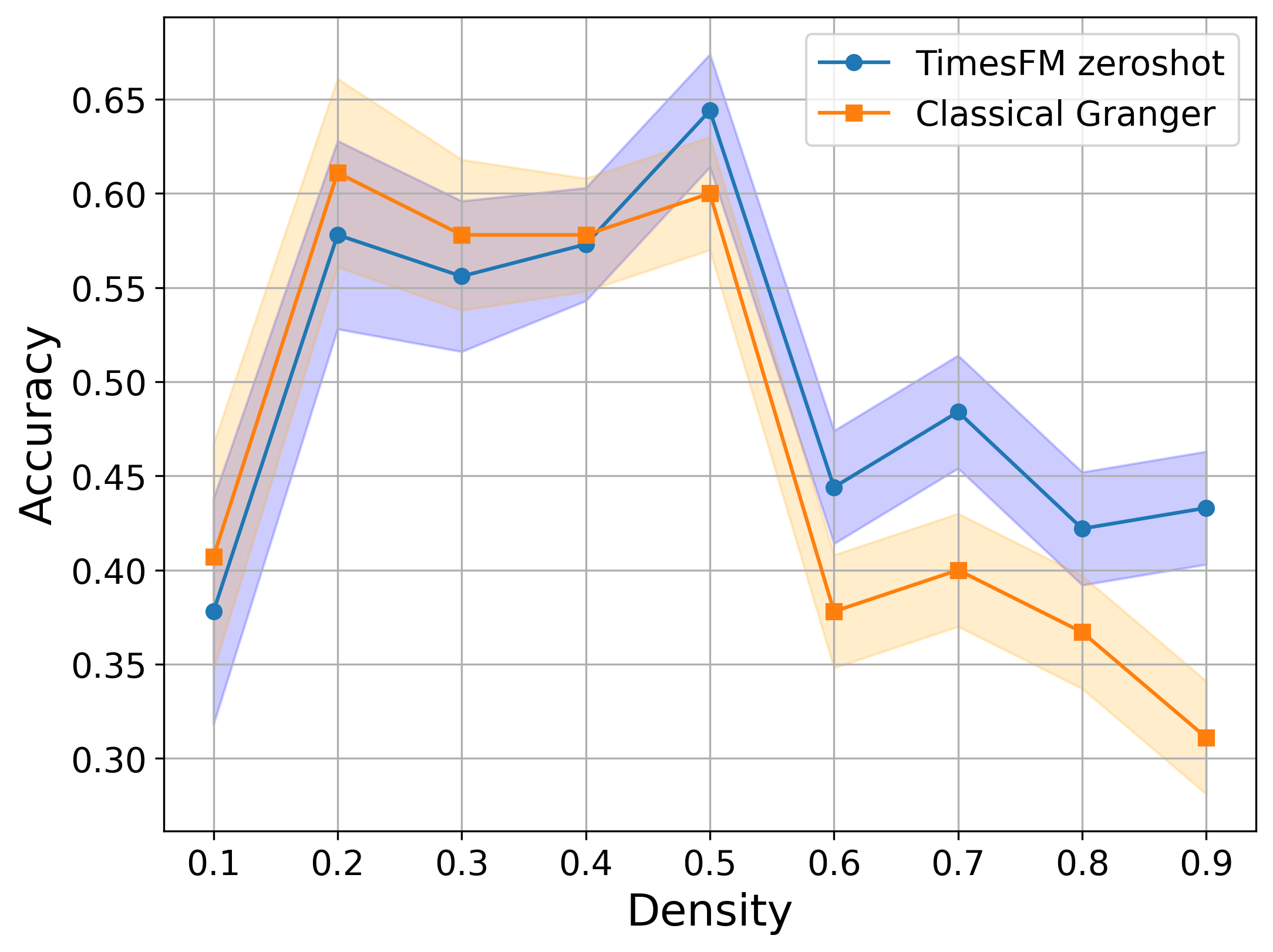}
    \caption{}
  \end{subfigure}
  \begin{subfigure}[b]{0.24\textwidth}
    \includegraphics[width=\linewidth]{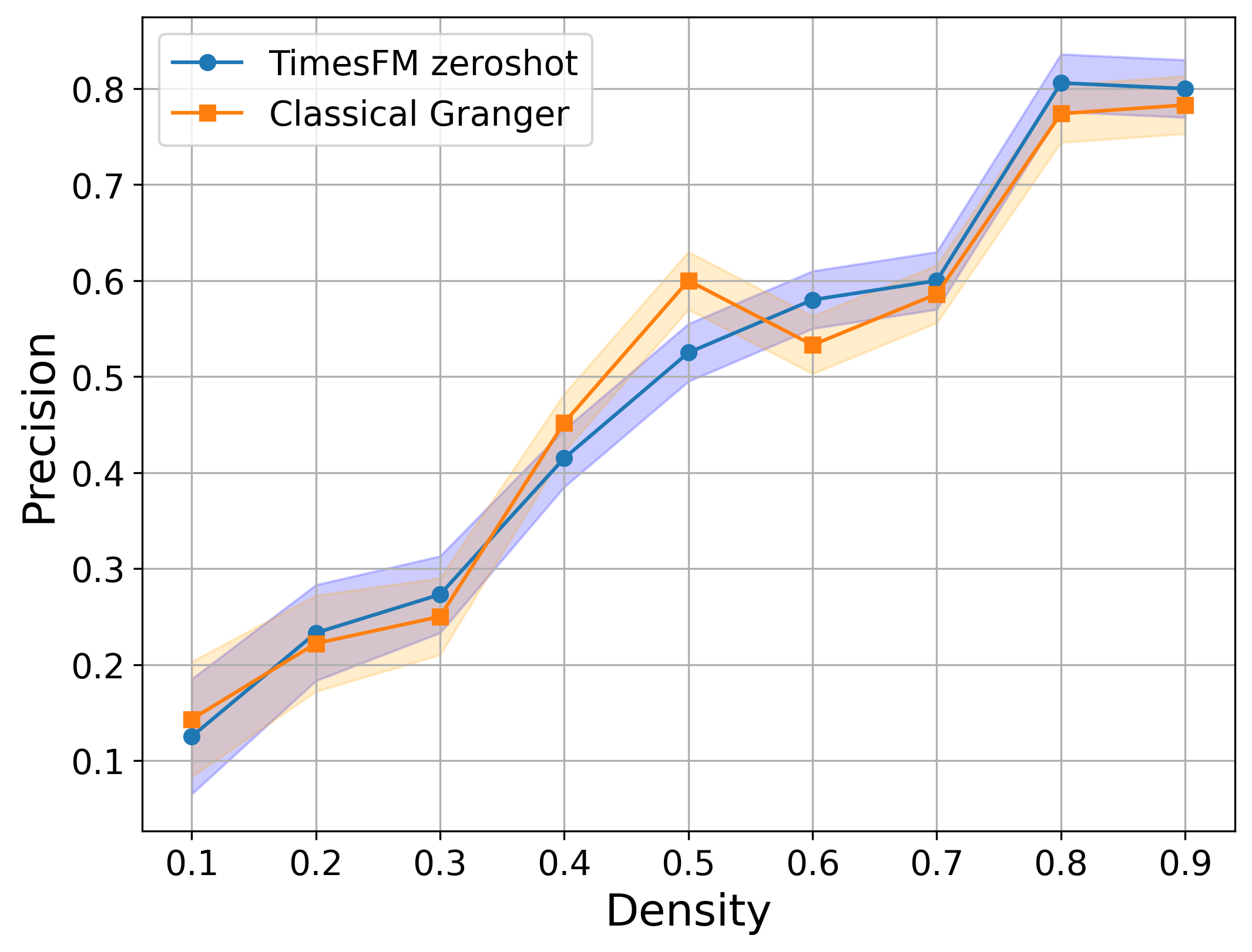}
    \caption{}
  \end{subfigure}
  \begin{subfigure}[b]{0.24\textwidth}
    \includegraphics[width=\linewidth]{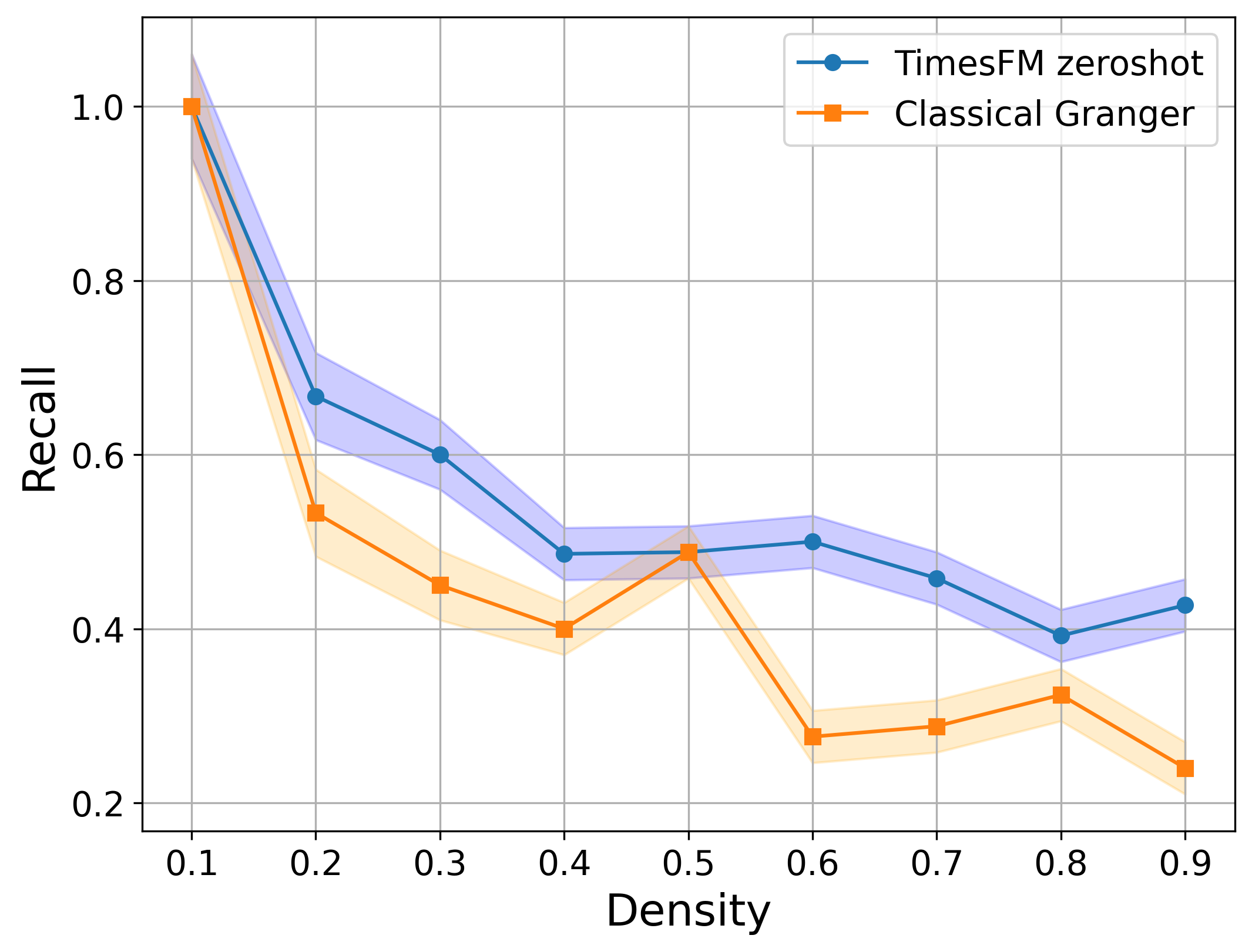}
    \caption{}
  \end{subfigure}
  \begin{subfigure}[b]{0.24\textwidth}
    \includegraphics[width=\linewidth]{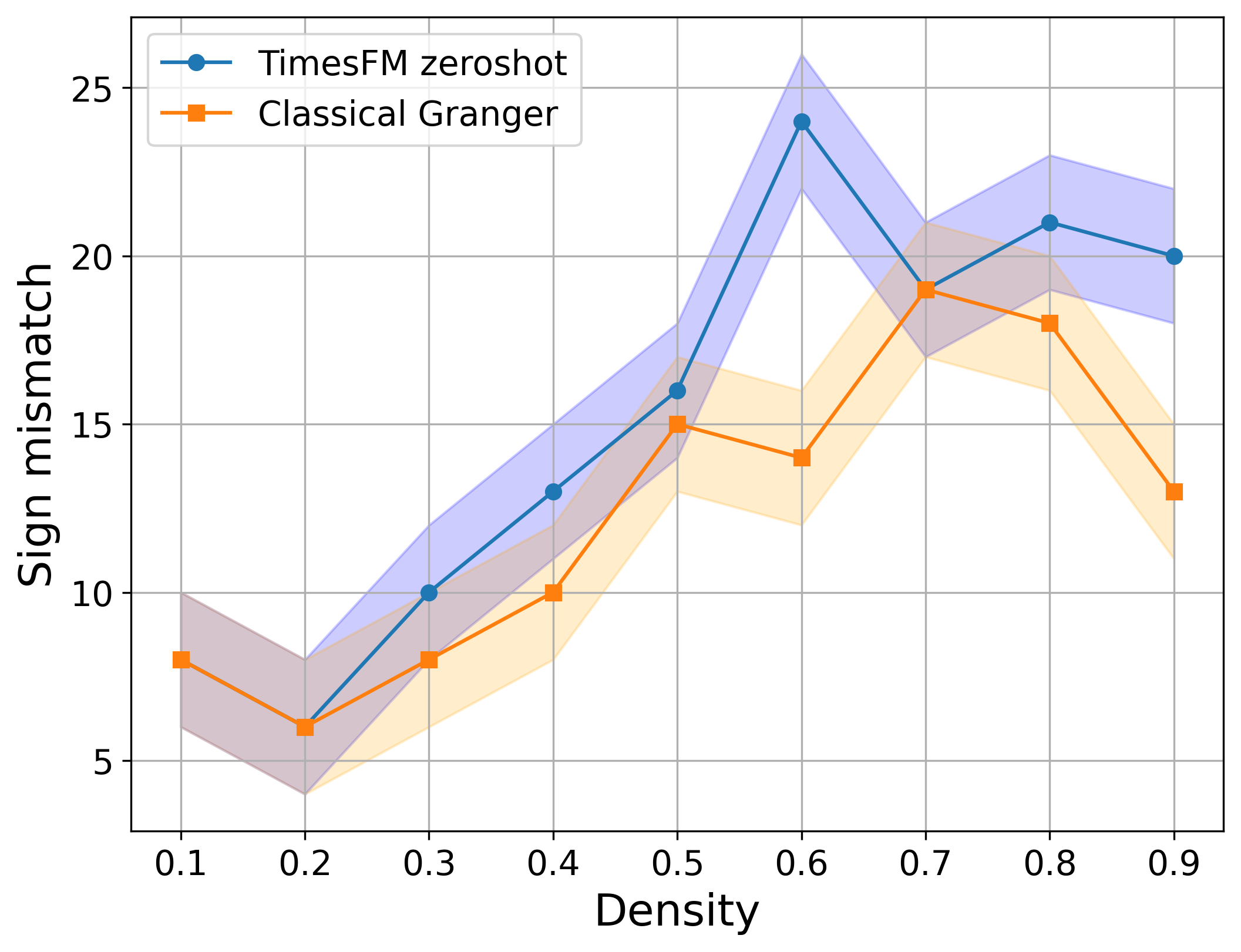}
    \caption{}
  \end{subfigure}
  \caption{(a) Accuracy, (b) Precision, (c) Recall, (d) Causality Sign mismatch for both methods varying the density (number of present causalities)  of cauality present in the networks with 10 nodes.}
  \label{fig:MOU}
\end{figure*}
\begin{table}[!ht]
\centering
\caption{Accuracy, Precision, and Recall for TimesFM-based and classical Granger causality for the 3-node networks.}
\begin{tabular}{lccc}
\hline
\textbf{Method} & \textbf{Metric} & \textbf{Mean} & \textbf{Variance} \\
\hline
\multirow{4}{*}{TimesFM zero-shot} 
& Accuracy  & 0.875  & 0.0016 \\
& Precision & 1.000  & 0.0000 \\
& Recall    & 0.750  & 0.0064 \\
\hline
\multirow{4}{*}{Granger } 
& Accuracy  & 0.875  & 0.0016 \\
& Precision & 0.8033 & 0.0026 \\
& Recall    & 1.000  & 0.0000 \\
\hline
\end{tabular}
\label{tab:3node}
\end{table}
Qualitatively, it was observed that the mismatch using Granger causality was very often caused by not detecting the causality, while for the foundation model approach, the mismatch was given by introducing a spurious causality between $X^{(1)}_t$ and $X^{(3)}_t$. 
For MOU-based networks, the results are shown in Figure \ref{fig:MOU}, 
where this behavior was even more pronounced, with qualitatively the mismatch given for Granger due to miss causality, and by TimesFM-based causality having more false positive. 
The metrics are calculated by varying the density of causality as described in Section 2.1.1.  While TimesFM does not require training time (zero-shot), its inference time is nearly 10× higher than ARIMA. Memory usage follows a similar pattern, with TimesFM requiring 2.1GB versus ARIMA's 350MB. Performing the analysis on the analysis on the human fMRI data gave a total mean 4650   and 3390 causal relationships using respectively the TimesFM and Classical Granger approach. However, without a ground truth it is not possible to validate the correctness.  Nevertheless, even with the fMRI data the causalities discovered using the foundation model are more than with the classical Granger approach. 

\begin{comment}
\begin{figure} 
\includegraphics[width=\linewidth]{OU_LLMvsGR.png}
\caption{Mismatched causality for the MOU process with varying density in data generation, disregarding causality sign.}
\label{fig:causality1}
\end{figure}

\begin{figure} 
\includegraphics[width=\linewidth]{foundation_vs_granger.png}
\caption{Mismatched causality for the MOU process with varying density in data generation, considering  causality sign.}
\label{fig:causality2}
\end{figure}
\end{comment}
\vspace{-0.5cm}
\section{Discussion}
\vspace{-0.2cm}
\label{sec:discussion}
Our results indicate that TimesFM, used in a zero-shot setting, consistently achieved the lowest reconstruction error in both control and patient datasets. This finding suggests that large foundation models trained on extensive time-series corpora can generalize effectively even to domains not explicitly represented during pretraining. However, statistical testing showed no significant differences in the zero-shot setting for controls using summarized data but only using the patients data; further analyses with ANOVA on region-level data will be conducted. Overall, the zero-shot performance of TimesFM was comparable to traditional methods. Interestingly, LR performed markedly worse in patients compared to controls. A plausible explanation is that patient time series exhibit stronger nonlinear patterns—such as irregular fluctuations and abrupt shifts driven by pathological mechanisms \cite{falco2024functional} — that violate linear models' assumptions. In contrast, control time series are comparatively more stationary and linear, allowing better performances in that group. This underscores the importance of model flexibility in analyzing pathological data. Fine-tuning TimesFM substantially improved performance, particularly in the patient subgroup; however, our focus here is to demonstrate   zero-shot capability. Linear Granger causality is constrained by its reliance on  vector autoregressions, limiting it to linear dependencies. In contrast, TimesFM, exploiting neural forecasting with attention and sequence modeling, can capture nonlinear interactions, time-varying dependencies, and long-range effects. For the simple 3-node case, no particular difference was observed, while improvements are visible for MOU experiments especially increasing the density (number of introduced causalities). 
Both methods perform modestly, as the task is challenging and the multiple test correction mitigate the effect.The foundation model is more accurate at higher causality densities. Its higher recall indicates fewer missed cases, but it also produces more sign mismatches, reflecting a greater tendency to detect causalities. In contrast, Granger causality yields more false negatives by missing true causalities. 
In summary, the foundation model allows more accurate time-series reconstruction and, consequently, more sensitive causal inference. However, not all detected influences necessarily reflect true causal relationships: some may arise from hidden confounders or shared drivers. As a result, the TimesFM approach exhibited a higher rate of false-positive causalities compared with standard Granger analysis.

\vspace{-0.3cm}
\section{Conclusion}
\vspace{-0.1cm}
%Overall, the transfer performance in- and out-of-distribution 
Our study suggests that even without fine-tuning, foundation models applied to time series can achieve reasonable performance in early event prediction for clinically relevant labels. However, causal discovery remains challenging, even when evaluated against synthetic datasets with known ground truth. Future work may explore incorporating sparsity-inducing approaches to mitigate false positives and improve the reliability of inferred causal relationships.

%\section*{Acknowledgment}
%This work was supported by NIH Grant XXXXX and NSF Award YYYYY. We thank the XYZ Center for Neuroimaging for data acquisition support.
\vspace{-0.2cm}
\section{RELATION TO PRIOR WORK}
\label{sec:prior} \vspace{-0.1cm}
Granger causality is a widely used tool to infer directional interactions from neural time series \cite{granger1980, brovelli2004, bressler2011}. % \cite{bayazi2024general}. % often require data-intensive end-to-end training. %We propose a residual-based method extending TimesFM to Granger-like causality without retraining, exploiting its ability to model nonlinear dependencies. 
Unlike \cite{chen2025large} and \cite{bayazi2024general}, which uses task-specific architectures, our approach leverages residuals of a pre-trained foundation model to also investigate causality, linking classical statistical tests with zero-shot causal inference in neuroscience.

%\vfill\pagebreak

\balance
\bibliographystyle{IEEEbib}
\bibliography{refs}

\end{document}